\documentclass{article}

\usepackage{microtype}
\usepackage{graphicx}
\usepackage{booktabs}
\usepackage{tikz}
\usepackage{pgfplots}
\usepackage{pgfplotstable}
\usepackage{subcaption}
\usepackage{amsmath}
\usepackage{amsthm}
\usepackage[normalem]{ulem}
\usepackage{hyperref}
\usepackage[utf8]{inputenc}

\usepackage[accepted]{icml2020}

\icmltitlerunning{Deep Representation Learning and Clustering of Traffic Scenarios}

\begin{document}

\twocolumn[
\icmltitle{Deep Representation Learning and Clustering of Traffic Scenarios}

\icmlsetsymbol{equal}{*}

\begin{icmlauthorlist}
\icmlauthor{Nick Harmening}{bmw}
\icmlauthor{Marin Biloš}{tum}
\icmlauthor{Stephan G\"unnemann}{tum}

\end{icmlauthorlist}

\icmlaffiliation{tum}{Technical University of Munich, Germany}
\icmlaffiliation{bmw}{BMW Group, Munich, Germany}

\icmlcorrespondingauthor{Nick Harmening}{nick.harmening@bmw.de}
\icmlcorrespondingauthor{Marin Biloš}{bilos@in.tum.de}
\icmlcorrespondingauthor{Stephan G\"unnemann}{guennemann@in.tum.de}

\icmlkeywords{Autonomous Driving, Deep Representation Learning, ICML}

\vskip 0.3in
]

\printAffiliationsAndNotice{}

\begin{abstract}
Determining the traffic scenario space is a major challenge for the homologation and coverage assessment of automated driving functions. In contrast to current approaches that are mainly scenario-based and rely on expert knowledge, we introduce two data driven autoencoding models that learn a latent representation of traffic scenes. First is a CNN based \textit{spatio-temporal} model that autoencodes a grid of traffic participants' positions. Secondly, we develop a pure \textit{temporal} RNN based model that auto-encodes a sequence of sets. To handle the unordered set data, we had to incorporate the permutation invariance property. Finally, we show how the latent scenario embeddings can be used for clustering traffic scenarios and similarity retrieval.

\end{abstract}

\section{Introduction}
\label{introduction}

\begin{figure}[t]
        	\centering
        	\begin{subfigure}[b]{0.22\textwidth}
        		\centering
        		\includegraphics[scale=0.6]{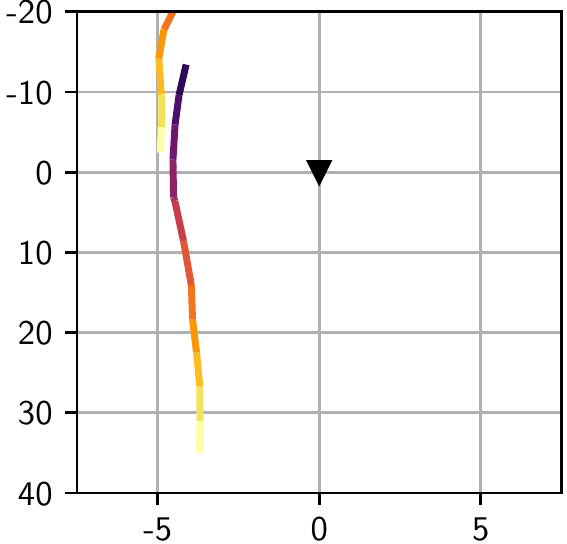}
        		\caption{Two trajectory scenario.}
        	\end{subfigure}
        	\begin{subfigure}[b]{0.22\textwidth}
        		\centering
        		\includegraphics[scale=0.6]{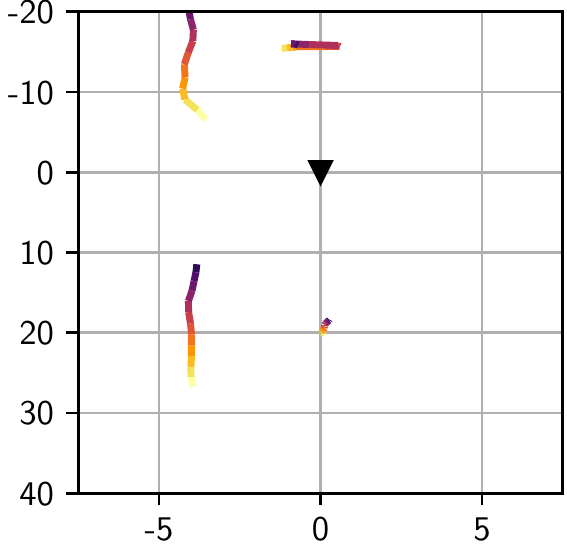}
        		\caption{Four trajectory scenario.}
        	\end{subfigure}
        \caption[Dense Traffic Retrieval Example]{Traffic scenarios with a fixed ego-vehicle that faces downwards (black triangle) and the relative movement of the traffic participants. The darker the color of the traffic participant's trajectory the later the position was taken in time.}\label{fig:scenarios}
    \end{figure}

While automated driving functions advance more and more, the question of validating them is still not answered. One way to measure a developed function's reliability is the percentage of successfully handled traffic scenarios (e.g.\ Overtaking-Scenario, Cut-In-Scenario, etc.) over all scenarios it will ever face.
Therefore, if we want to estimate how well the function covers the real-world distribution of the scenarios, we need to develop a way to compare these distributions or cluster the scenarios. This way we can make assessments on specific groups of scenarios. In our work, a scenario consists of the list of relative traffic-participant trajectories (see Figure \ref{fig:scenarios}).    In industry, current concepts for finding clusters of scenarios are  largely based on expert knowledge and approach the task by defining and parametrizing a discrete scenario-catalogue.
This bottom-up approach has obvious limitations in the number and complexity of scenarios that can be defined. It is also questionable, if a full coverage of scenes can ever be reached.

In this paper we introduce two deep representation learning methods (the \textit{Spatio-temporal Autoencoder} and the \textit{Sequential Deep Set Prediction Network}) that learn a latent space of traffic scenarios by the principle of auto-encoding. The main idea is to process a scenario with an encoder to get a latent scenario representation and try to reconstruct it with a decoder. This latent representation will later be used for scenario clustering. The main challenges for processing a scenario in a neural network are: 1) the traffic participants in the scene do not have an inherent ordering, 2) the number of traffic participants can vary, 3) the length of the traffic participants trajectories in a scenario can vary. With the proposed methods we overcome exactly these issues.

\section{Related Work}
\label{related_work}
  While there already exists a huge body of research  focusing on clustering trajectories \cite{Trajectory_Clustering}, these methods only compare single trajectories. Most of them are based on classical distance functions between trajectories combined with clustering methods, e.g.\ DB-SCAN. A deep learning based approach is provided by \citet{DeepRepresentationLearning}, who use a sequence to sequence autoencoder to learn representations of trajectories. The representations are afterwards clustered. Since a scenario can be comprised of several traffic participants and therefore several trajectories, these approaches cannot be directly applied to scenario clustering.

    \citet{DrivingEncounterScenarios} focus on clustering encountering scenarios consisting of two vehicles. They use the Dynamic-Time-Warping Matrix to encode the encountering scenario and use a convolutional autoencoder afterwards to learn a representation that can be clustered. This approach, unlike ours, does not scale to more than two trajectories.

    \citet{BEV_Paper} introduce a spatio-temporal data representation for classifying traffic scenarios. The data shows a top view on the traffic scene and encodes the location and the movement of the participants into a discrete, fixed sized grid. The basic grid consists of an occupancy feature map, and lateral and longitudinal velocity of the traffic participants. To encode time information the authors introduce a shading of last positions in the occupancy grid or they add extra channels with a previous frame. Overall, this results in a three-dimensional grid (two spatial and one channel dimension) to represent the scenario. They run a 2D CNN over the grid data to predict classes as \textit{Ego vehicle speed lager than 1m/s, leading vehicle ahead, other vehicles overtaking, etc}. In contrast, our spatio-temporal model extends this by having time as an additional input dimension. This allows us to use a 3D CNN for grid auto-encoding.

\section{Models}\label{methodology}
  A scenario is defined with a set of trajectories over a certain number of time frames. Each trajectory describes the movement of one traffic participant relative to the ego-vehicle (the drivers vehicle).

    The two models we propose differ drastically due to the different data representation that is used for each of them. The traffic participants in the \textit{spatio-temporal data representation} are ordered by their spatial position into a grid, which already solves the problem of permutations in the list of trajectories. The \textit{sequence of sets data representation} is an un-ordered and purely temporal representation, which does not need to be discretized into a fixed spatial grid. The model applied to this representation was developed while paying attention to permutation invariance.

    \subsection{Spatio-temporal Autoencoder}
\begin{figure}[t]
                	\centering
                	\includegraphics[scale=0.3]{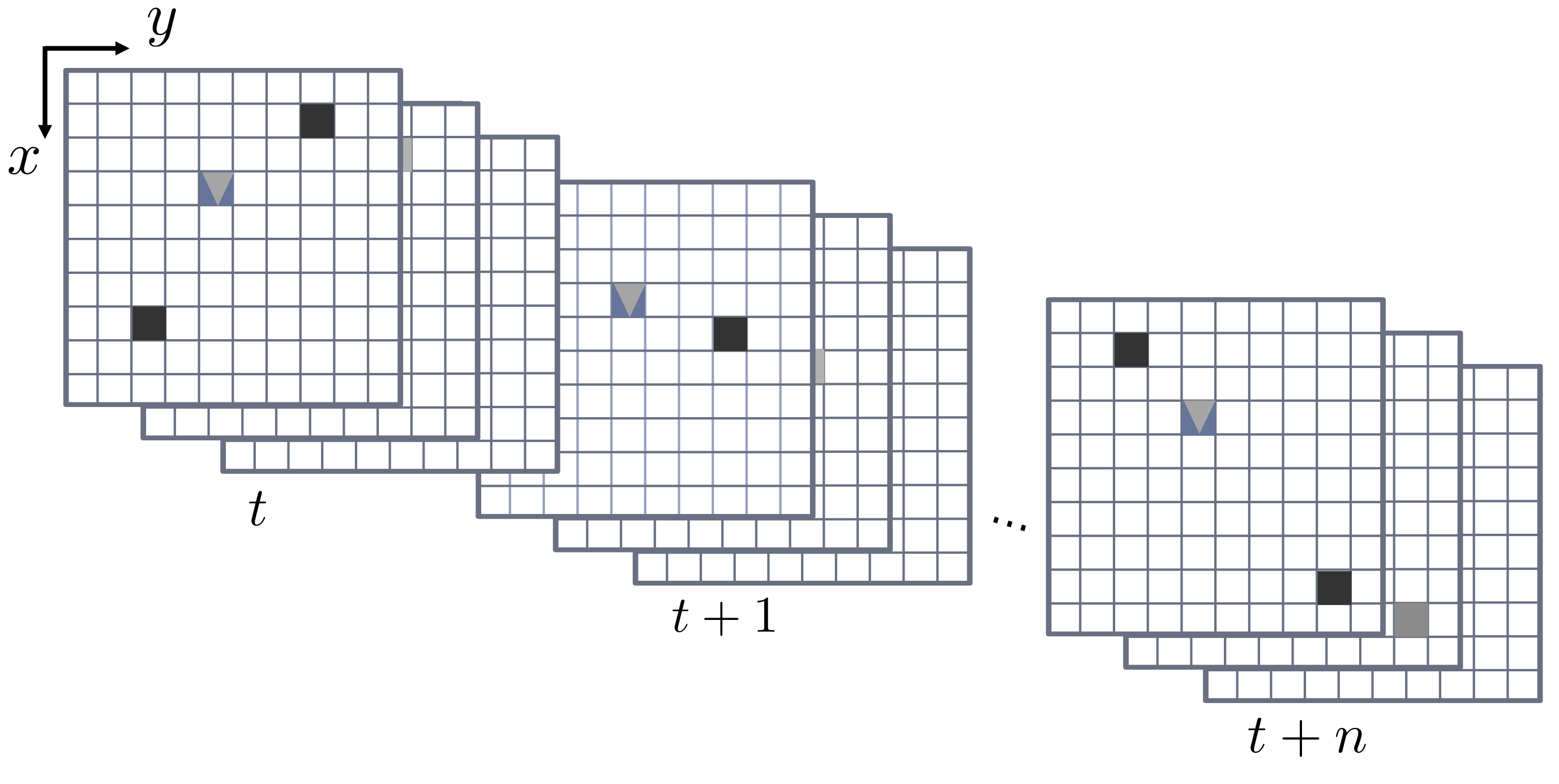}
                	\caption[Spatio-temporal grid]{Spatio-temporal grid $\mathbf{X}$ $\in {\rm I\!R}^{  d_t \times d_x \times d_y \times n_c}$ that covers $d_t$ frames from $t$ to $t+n$. $d_{x/y}$ describe the spatial dimensions of the grid and $n_c$ describes the number of channels.}\label{fig:stacked_bev_grid}
                \end{figure}

        Based on the 3D Stacked Velocity Grid \cite{BEV_Paper} we developed an extended grid representation to better capture temporal features (see Figure \ref{fig:stacked_bev_grid}). Instead of limiting us to two spatial dimensions and the channel dimension, the time axis is introduced, along which the frames are aligned. Each frame in a scene has one binary occupancy channel which encodes the traffic participants location in a scenario. The ego-vehicle is fixed to a central grid cell (fixed lateral and longitudinal offset) throughout all scenarios, therefore the traffic participants movements is always relative to the ego-vehicle. We can add additional channels that encode further features at the corresponding location, e.g.\ lateral and longitudinal velocity/acceleration, yaw-angle etc. The elegance of the spatio-temporal representation approach is that it can inherently deal with the issue of ordering, different trajectory lengths and a variable number of objects.

            To encode the four dimensional input grid $\mathbf{X}$, three dimensional convolutions which convolve over two spatial dimensions plus the time dimension are used. Three-dimensional CNNs are already used by e.g.\ \citet{Video_Classification} for large scale video classification and by \citet{Video_Segmentation} to perform video segmentation. After each convolution in the encoder, the batch-norm and max-pooling layer is applied. The decoder makes use of the up-convolutions and un-pooling layers, introduced in the DeconvNet \cite{DeconvNet} to decode the latent representation back to the input dimensionality. The resulting bottleneck vector $\mathbf{z}$ encodes the scenario in the latent representation and will be used for the clustering task later. Since velocity and acceleration are continuous, we use the mean squared error between the target grid $\mathbf{X}$ and the reconstructed grid $\hat{\mathbf{X}}$ as a reconstruction loss:

        \begin{equation}
        		\mathcal{L} (\mathbf{X},\hat{\mathbf{X}})= \sum_{a = d_t} \sum_{b = d_x} \sum_{c = d_y} \sum_{d = n_c} (x_{a,b,c,d} - \hat{x}_{a,b,c,d})^2
        \end{equation}

    \subsection{Sequential Deep Set Prediction Network}
        For the Sequential Deep Set Prediction Network (SeqDSPN) the list of trajectories is transformed into a sequence of sets $\mathbf{X} = [\mathbf{X}_1,\dots, \mathbf{X}_t,\dots,\mathbf{X}_T]$ that describe the scenario. Each frame of the scenario corresponds to a set of $n$ traffic participants that have the features such as position, velocity, etc., at the specific time-point $\mathbf{X}_t = \{(x_1,y_1,v_1)_t, \dots, (x_{n}, y_{n}, v_{n})_t\}$.

\begin{figure}[t]
   			\centering
   			\includegraphics[scale=0.4]{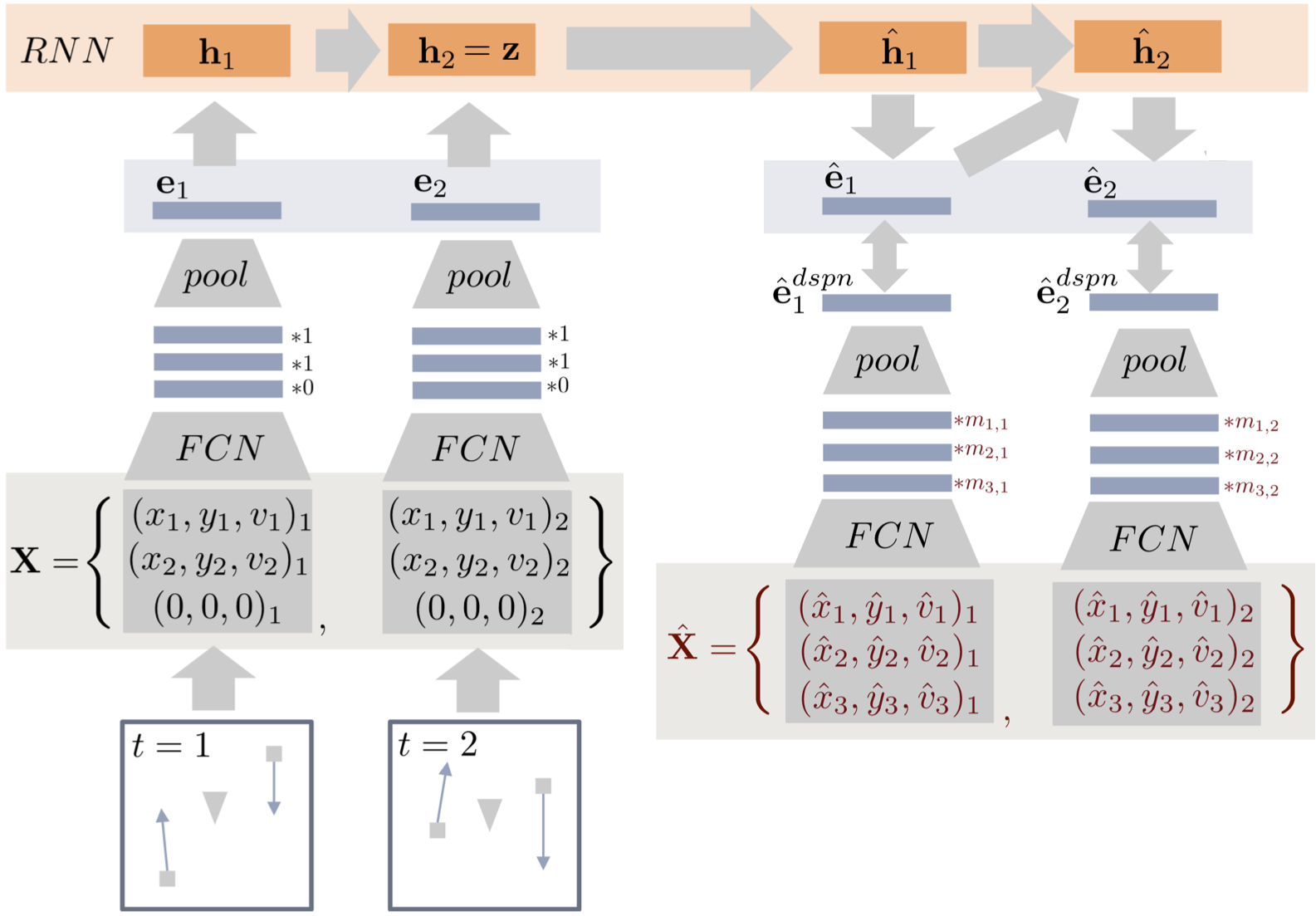}
   			\caption[Sequential Deep Set Prediction Network]{SeqDSPN architecture for two frames. The red variables are randomly initialized and optimized by the inner optimization loop of the DSPN. The goal of the inner optimization is to bring $\hat{\mathbf{e}}_t^{dspn}$ close to  $\hat{\mathbf{e}}_t$. }\label{fig:sdspn}
       	\end{figure}

        We first encode each set $\mathbf{X}_t$ into an embedding vector $\mathbf{e}_t$ with a permutation invariant network, specifically a deep set network \cite{DeepSets}. First, fully connected layers compute an embedding for each element of the set, then a feature-wise max-pooling operation on these embeddings returns a permutation invariant set embedding. While the number of traffic participants can vary, the final set embedding always has the same dimensionality. This way we encode the main features of one specific frame (see Figure~\ref{fig:sdspn}).

        The second step is auto-encoding the sequence of embeddings in a sequence to sequence fashion \cite{Seq2Seq}. The final hidden vector of the encoding RNN yields the scenario representation $\mathbf{z}$ and will be used for the clustering task later.

        The final step is the reconstruction of the sets based on the reconstructed embedding vectors $\hat{\mathbf{e}}_t$. For this purpose we use the Deep Set Prediction Network \cite{Deep_Set_Prediction_Networks}, which can be seen as a predictor of sets, based on an embedding vector. The DSPN searches for a set that has a similar encoding to the given embedding vector. This set is then the predicted set. The working principle of the DSPN is to randomly initialize the set prediction $\hat{\mathbf{X}}_t$ and a mask. The mask is needed for predicting different set sizes. It allows the network to encode whether a certain set entry is actually intended to be predicted or if it should be omitted. After the random initialization, the values are optimized iteratively in an inner optimization loop. The inner optimization procedure is passing the set  $\hat{\mathbf{X}}_t$ and the corresponding mask through an encoder, yielding one embedding $\hat{\mathbf{e}}_t^{dspn}$. The mean squared error between this embedding and input embedding $\hat{\mathbf{e}}_t$ is then used as the loss function we optimize for in the inner loop. After this inner optimization, $\hat{\mathbf{e}}_t^{dspn}$ optimally converged to $\hat{\mathbf{e}}_t$. All together these three steps enable the auto-encoding of the sequence of sets.

        For the loss, \citet{Deep_Set_Prediction_Networks} propose to use the Chamfer's Distance $O(n^2)$ or the Hungarian Algorithm $O(n^3)$ to determine the distance  $\mathcal{L}_{set}$ between the input sequence of sets $\mathbf{X}$ and the reconstructed ones $\hat{\mathbf{X}}$, which is additionally multiplied with the predicted mask. Chamfer's Distance, for example, matches up the closest instances from  $\mathbf{X}$ in $\hat{\mathbf{X}}$ and vice versa:
		\begin{equation}
        \begin{aligned}
        	\mathcal{L}_{set}(\hat{\mathbf{X}}, \mathbf{X})  &=
        	\sum_{i} \min_j ||\hat{\mathbf{x}}_i-\mathbf{x}_j||^2 \\
        	 &+ \sum_{j} \min_i || \mathbf{\hat{x}}_i-\mathbf{x}_j||^2
        \end{aligned}
        \end{equation}
		To further stabilize the training process of the RNN, we add the mean squared error between the embeddings and the reconstructed embedding, which yields the final loss:
        \begin{equation}
	   		\mathcal{L}_{dspn} = 	\mathcal{L}_{set} + \lambda\sum_{t=1}^{T}\mathcal{L}_{mse}(\mathbf{e}_t,\hat{\mathbf{e}}_t)
   		\end{equation}
   		where $\lambda$ is a weighting parameter. Additional hyperparameter in the SeqDSPN are the inner learning rate and the number of inner optimization steps of the DSPN.

\section{Experimental Results}\label{experimental_results}
    In this experimental section the two models are applied to a sub-set of the \textit{BMW Crowd Data Collection}. The data is collected by vehicles with automated driving functions that send the high level environmental model, i.e.\ positions and velocities of the traffic participants on the highway. For our purpose we selected a subset of the data and randomly cut it into five second scenes and dropped all scenes where no traffic participant was present. The train-set and validation-set consists of 42,000 scenarios (80/20 split). The held out test-set was recorded one month later consisting of 90,000 five second samples.

    For the Spatio-temporal AE we used a grid with the spatial dimensions of 30 $\times$ 30, which (after squeezing laterally and stretching longitudinally) results in a coverage of 15 meters laterally and 60 meters in the longitudinal direction. Having 13 time-frames (2.5 Hz) and two channels (occupancy and longitudinal velocity) finally yield the input matrix $\mathbf{X} \in {\rm I\!R}^{ 13 \times 30 \times 30 \times 2}$. The occupancy channels of the target grid is additionally smoothed by a Gaussian kernel to avoid sparsity. The bottleneck size is determined to 64. In the SeqDSPN we use a two layer LSTM with the hidden size 64 to have the same size as in the Spatio-temporal AE. A detailed description of the architectures can be found in the Appendix, see Table \ref{tab:bev_architectures} and Table \ref{tab:dspn_architectures}. Although the SeqDSPN can handle a varying set size $n$, we pad the input sets to the maximum set size to enable GPU acceleration. If there are more traffic participants present, only the $n$ closest ones are considered. The reconstruction of sets includes a mask where the network can encode its prediction for the set-size.

    \begin{figure}[htpb]
        \centering
        \input{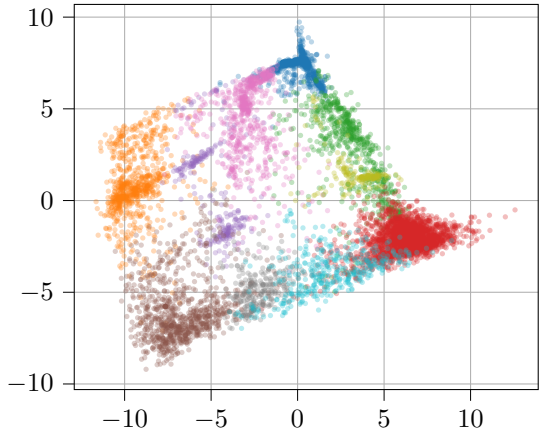}
        \caption{Downprojected (via PCA), clustered latent validation space of the Spatio-temporal AE.}
        \label{fig:clustered_latent_space_BEV}
    \end{figure}

    Figure \ref{fig:clustered_latent_space_BEV} shows the latent space of the validation set of the Spatio-temporal AE, downprojected by PCA. The colour indicate clusters that we found via hierarchical clustering \cite{Murphy}. The four clusters constructing the edges of the latent space can semantically be described as \textit{Overtaking} (red), \textit{Being Overtaken} (orange), \textit{Dense Traffic} (brown) and \textit{Almost Empty} (blue). Analyzing the nearest neighbours in this latent space indicates a very strong smoothness of the learned representations (see Figure \ref{fig:retrieval_dense} in the Appendix). One should note that finding similar scenarios usually includes very complex and custom similarity functions that work on the environmental model. By using the euclidean distance in the latent space, scenario retrieval is enabled without any customized distance function.

\begin{figure}[htpb]
        \centering
        \input{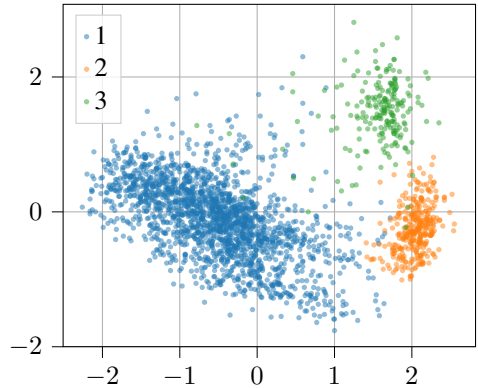}
        \caption{Auto-labeled latent test-space of the SeqDSPN, reduced via PCA. This particular latent space yielded a V-Measure of 0.92 after clustering. The analog plot for the Spatio-temporal AE can be found in the Appendix, Figure \ref{fig:labeled_latent-space_BEV}.}
        \label{fig:labeled_latent-space}
    \end{figure}

    For quantifying the models performance in terms of semantically correct separation of scenarios, we implemented an auto-labeler for three classes (Class 1: Ego Overtakes, Class 2: Leading vehicle ahead, Class 3: Ego is being overtaken). These automatically generated labels are used to evaluate the clustering quality of the latent space that was learned in an unsupervised fashion. The simple label logic for the three scenarios was applied to the test set. From the initial test dataset of 90,000 scenes, around 19,500 were labeled (Class 1: 15,000, Class 2: 3,000, Class 3: 1,500). To compare the two methods, we train until convergence six times per approach. After training, the trained models were used to compute the latent representation space of the labeled test-set (see Figure \ref{fig:labeled_latent-space}).

    The latent space is then clustered via hierarchical clustering with a number of clusters equal to three. The cluster to class assignment is done by a majority vote of the instances that are contained in the cluster. Finally, the clustering performance is evaluated by the V-measure \cite{v_measure}. We report the mean and standard deviation for the runs in Table \ref{sample-table}.

    \begin{table}[htpb]
    \caption{Clustering Performance averaged over six runs: $\mu \pm  \sigma$.}
    \label{sample-table}
    \vskip 0.15in
    \begin{center}
    		\begin{tabular}{ c |  c }
    			Model Name  & $V_1$-Measure \\
    			\hline  \hline
    			SeqDSPN & 0.77 $\pm$ 0.17 \\
    			\hline
    			BEV AE & 0.76 $\pm$ 0.18 \\
    		\end{tabular}
    \end{center}
    \vskip -0.1in
    \end{table}

    The numbers indicate that approaches performed equally well in this setup. One way the high variance could be explained is due to the uneven distribution of the classes. Since Class 1 represents 75$\%$ of the labels,  it is usually clustered well. Whereas Class 2 (16 $\%$) and Class 3 (9 $\%$) sometimes happen to be clustered into the same group. Even though a better tailored clustering algorithm might have increased the scores, we deliberately selected the very robust hierarchical clustering to avoid an overfitting to latent space characteristics of a certain model.

\section{Conclusion}\label{conclusios}
 In this paper we introduced two unsupervised deep learning approaches that proved to be capable of learning an expressive latent space of traffic scenarios for a real-world highway dataset. The latent space is the first step towards a data driven understanding scenario distributions. Additionally, we showed that the latent space can be used for retrieving similar scenes from a dataset.

In our work we focused on the environmental data representations via sets and grids. Further study of using temporal graphs as a data representation could be of interest. In addition, investigation on how to better cut drives to scenarios is also required (e.g. event-based cutting). The last research direction for future work are generative models for traffic scenarios. Based on the auto-encoding architecture of our methods, switching to a variational autoencoder would already be a starting point.

\section*{Acknowledgements}
This research was made possible by the financial support and the data provision of the BMW Group. We are particularly grateful to Tobias Freudling and Marc Neumann for sharing their extensive knowledge in the area of Autonomous Driving and for the productive discussions on the application of the developed methods.

\bibliography{bibliography}
\bibliographystyle{icml2020}

\appendix

\newpage
\section{Supplementary}\label{supplementary}

\begin{figure}[htpb]
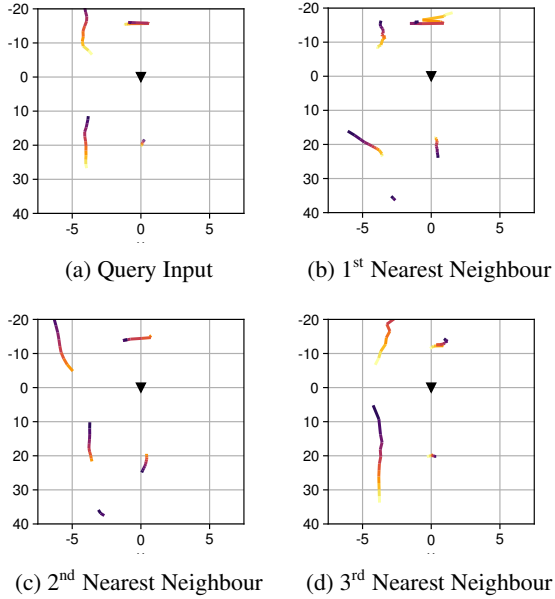

    	\centering
    	\begin{subfigure}[b]{0.22\textwidth}
    		\centering
    		\resizebox{.95\linewidth}{!}{\input{1_dense.pgf} }
    		\caption{Query Input}
    	\end{subfigure}
    	\begin{subfigure}[b]{0.22\textwidth}
    		\centering
    		\resizebox{.95\linewidth}{!}{\input{2_dense.pgf} }
    		\caption{1\textsuperscript{st} Nearest Neighbour}
    	\end{subfigure}
    	\begin{subfigure}[b]{0.22\textwidth}
    		\centering
    		\resizebox{.95\linewidth}{!}{\input{3_dense.pgf} }
    		\caption{2\textsuperscript{nd} Nearest Neighbour}
    	\end{subfigure}
    	\begin{subfigure}[b]{0.22\textwidth}
    		\centering
    		\resizebox{.95\linewidth}{!}{\input{4_dense.pgf} }
    		\caption{3\textsuperscript{rd} Nearest Neighbour}
    	\end{subfigure}\caption[Dense Traffic Retrieval Example]{Dense Traffic Retrieval examples. The darker the colour of the trajectory the later the position was taken in time. The nearest neighbours were computed in the latent space of the test-set computed by the Spatio-temporal AE.}\label{fig:retrieval_dense}
\end{figure}

\begin{figure}[htpb]
    \centering
    \input{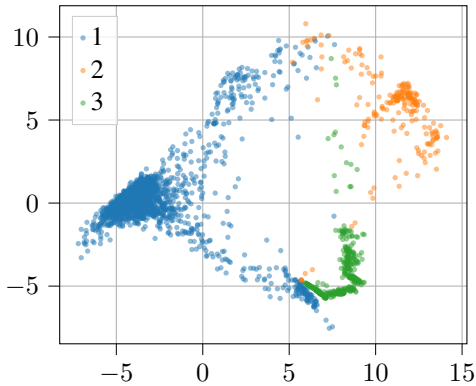}
    \caption{Auto-labeled latent test-space of the Spatio-temporal AE, reduced via PCA. This particular latent space yielded a V-Measure of 0.91 after clustering.}
    \label{fig:labeled_latent-space_BEV}
\end{figure}

\begin{table}[htbp]
	\centering
	\begin{tabular}{  c |  c }
		Parameter  & Value \\
		\hline  \hline
		\textbf{Layer 1} &\\
		Conv. filter size &  $\left(5, 7, 7\right)$ \\
		No. filters & 4\\
		Conv. stride&  1  \\
		Pooling mask size  &  $\left(2, 2, 2\right)$ \\
		Pooling mask stride  &  $\left(2, 2, 2\right)$ \\
		Batch norm &  yes \\
		\hline
		\textbf{Layer 2} &\\
		Conv. filter size&  $\left(3, 5, 5\right)$ \\
		No. filters & 6\\
		Conv. stride & 1  \\
		Pooling mask size  &  $\left(2, 2, 2\right)$ \\
		Pooling mask stride &  $\left(2, 2, 2\right)$ \\
		Batch norm & yes \\
		\hline
		\textbf{Layer 3} &\\
		Conv. filter size &  $\left(3, 3, 3\right)$ \\
		No. filters &  8\\
		Conv. stride &  1  \\
		Pooling mask size  &  $\left(1, 2, 2\right)$ \\
		Pooling mask stride & $\left(1, 2, 2\right)$ \\
		Batch norm & yes \\
		\hline
		Bottleneck size & 64\\
		Gaussian Kernel & 5 $\times$ 5 ,  $\sigma$= 1\\
		Loss &  MSE
	\end{tabular}\caption[BEV AE Architectures]{Spatio-temporal AE architecture}\label{tab:bev_architectures}
\end{table}

\begin{table}[htbp]
	\centering
	\begin{tabular}{  c | c }
		Parameter  & Value \\
		\hline  \hline
		\textbf{Encoder} &\\
		No. Neurons 1 Layer &  8 \\
		No. Neurons 2 Layer &  32 \\
		\hline
		\textbf{LSTM} &\\
		Hidden Size &  64 \\
		No. Layers & 2 \\
		\hline
		Maximum set size &  3 \\
		Bottleneck size &  64 \\
		Inner optimization steps &  25 \\
		Loss &   Chamfer's
	\end{tabular}\caption[SeqDSPN Architectures]{SeqDSPN architecture}\label{tab:dspn_architectures}
\end{table}

\end{document}